\documentclass[conference]{IEEEtran}
\IEEEoverridecommandlockouts
\usepackage{cite}
\usepackage{amsmath,amssymb,amsfonts}
\usepackage{multicol}
\usepackage{algorithmic}
\usepackage{multirow}
\usepackage{enumerate}
\usepackage{etoolbox}
\usepackage{amsmath,graphicx}
\usepackage{lipsum}
\usepackage{bbm}
\usepackage{dsfont}
\usepackage{yfonts}
\usepackage{amssymb}
\usepackage{graphicx}
\usepackage{mathrsfs}
\usepackage{textcomp}
\usepackage{xcolor}
\def\BibTeX{{\rm B\kern-.05em{\sc i\kern-.025em b}\kern-.08em
    T\kern-.1667em\lower.7ex\hbox{E}\kern-.125emX}}
\begin{document}

\title{RWN: Robust Watermarking Network for Image Cropping Localization\\
\thanks{This work is supported by National Natural Science Foundation of China under Grant U20B2051, U1936214. Qichao Ying and Xiaoxiao Hu contribute equally. $^{\star}$ Corresponding author: Zhenxing Qian (zxqian@fudan.edu.cn).
We are grateful to the anonymous reviewers for their constructive and insightful suggestions on improving this paper.}
}

\author{\IEEEauthorblockN{Qichao Ying}
\IEEEauthorblockA{\textit{School of Computer Science} \\
\textit{Fudan University, Shanghai, China}\\
shinydotcom@163.com}
\and
\IEEEauthorblockN{Xiaoxiao Hu}
\IEEEauthorblockA{\textit{School of Computer Science} \\
\textit{Fudan University, Shanghai, China}\\
xxhu21@fudan.edu.cn}
\and
\IEEEauthorblockN{Xiangyu Zhang}
\IEEEauthorblockA{\textit{School of Computer Science} \\
\textit{Fudan University, Shanghai, China} \\
xyzhang20@fudan.edu.cn}
\and
\IEEEauthorblockN{Sheng Li}
\IEEEauthorblockA{\textit{School of Computer Science} \\
\textit{Fudan University, Shanghai, China} \\
lisheng@fudan.edu.cn}
\and
\IEEEauthorblockN{Zhenxing Qian$^{\star}$}
\IEEEauthorblockA{\textit{School of Computer Science} \\
\textit{Fudan University, Shanghai, China} \\
zxqian@fudan.edu.cn}
\and
\IEEEauthorblockN{Xinpeng Zhang}
\IEEEauthorblockA{\textit{School of Computer Science} \\
\textit{Fudan University, Shanghai, China} \\
zhangxinpeng@fudan.edu.cn}
}

\maketitle

\begin{abstract}
Image cropping can be maliciously used to manipulate the layout of an image and alter the underlying meaning.
Previous image crop detection schemes only predicts whether an image has been cropped, ignoring which part of the image is cropped. 
This paper presents a novel robust watermarking network (RWN) for image crop localization. We train an anti-crop processor (ACP) that embeds a watermark into a target image. The visually indistinguishable protected image is then posted on the social network instead of the original image. 
At the recipient’s side, ACP extracts the watermark from the attacked image, and we conduct feature matching on the original and extracted watermark to locate the position of the crop in the original image plane. 
We further extend our scheme to detect tampering attack on the attacked image. Besides, we explore a simple yet efficient method (JPEG-Mixup) to improve the generalization of JPEG robustness. According to our comprehensive experiments, RWN is the first to provide high-accuracy and robust image crop localization. Besides, the accuracy of tamper detection is comparable with many state-of-the-art passive-based methods.
\end{abstract}


\section{Introduction}
\noindent Image manipulation leads to severe security threats, where misleading photojournalism, copyright violation or fabricating stories can be a means for some writers or politicians to potentially influence public opinion. Researchers have developed a number of schemes to detect various kinds of digital attacks, e.g., tampering~\cite{wu2019mantra,kwon2021cat,dong2021mvss}, DeepFake~\cite{li2020face} and cropping~\cite{fanfani2020vision,yerushalmy2011digital,li2009passive,van2020dissecting}. 
Among them, cropping is an extremely cheap and effective way to manipulate the layout of an image and alter the underlying meaning. 
Existing crop detection algorithms~\cite{fanfani2020vision,yerushalmy2011digital} mainly focus on predicting whether an image is cropped. 
They are represented by detecting the exposing evidences of asymmetrical image cropping, e.g., the shift of the image center \cite{fanfani2020vision}, and the inconsistency of JPEG blocking artifacts \cite{yerushalmy2011digital,li2009passive}.
Despite the effectiveness of these works, different cropping behaviors in the real world usually lead to varied intentions. Crop detection is only a binary classification from this perspective, unable to distinguish benign crop behaviors from those malicious attacks such as discarding a visible watermark or removing a person.
Van et al.~\cite{van2020dissecting} proposes an image crop localization scheme to study how image cropping causes chromatic and vignetting aberration. But the tiny traces only exists in high-definition images and can be easily destroyed by image processing, which makes the scheme hard to be applicable.
Experiments show that the scheme cannot be applied on highly-compressed or low-quality images. Therefore, image cropping localization remains a big issue.

Watermarking~\cite{ying2019robust,zhu2018hidden, mun2019finding} aims at hiding information imperceptibly into the host data for covert purposes.
The technology focuses on robustness against possible digital attacks, such as image compression, noise adding, filtering, etc. Watermarking has been widely used in copyright protection and content authentication of images in multimedia. 
Recently, many novel watermarking schemes for image protection is proposed. 
Khachaturov el al.~\cite{khachaturov2021markpainting} proposes a watermarking-like adversarial method that prevents inpainting systems to perform normally on a protected image. In~\cite{yin2018deep}, the authors propose an adversarial attack method to cause errors in super-resolution model, including making the SR image has undesired style, be incorrectly classified in classification task, or generates wrong words in image caption task. The myth behind these schemes is that once the secret information successfully travel through the Online Social Network (OSN), the recipient can decode the information to administrate a number of tasks.

This paper explores the potential of robust watermarking on image cropping localization. We propose a robust watermarking network (RWN) for image crop localization by protecting the original image through watermarking. If the protected image is cropped, RWN is expected to locate the cropped region at the recipient's side.
Watermarking is utilized to hide imperceptible traces for cropping localization which is resilient against common attacks, namely, JPEG compression, etc. 
We further introduce a tamper detector to deal with tampering attack. 
Therefore, compared to previous works~\cite{van2020dissecting,fanfani2020vision}, the tough goal of finding natural-yet-fragile traces for image cropping can be circumvented by maintaining the overall image quality after watermark embedding.


We train an anti-crop processor (ACP) based on invertible neural network~\cite{kingma2018glow} to embed a watermark into a original image. The watermark is shared with the recipient. ACP produces a visually indistinguishable protected image, which is then posted on the social network instead of the unprotected version. On receiving the attacked version of the protected image, we use ACP again to extract the watermark, and conduct an efficient and classical feature matching algorithms (SURF~\cite{bay2006surf}) between the original and extracted watermark to 
determine where the crop was positioned in the original image plane. 
We further use the tamper detector to predict the tamper mask, and the features within the predicted tamper mask is discarded to prevent mismatching. 
Besides, we explore a simple yet efficient method (JPEG-Mixup) to improve the generalization of JPEG robustness.
We test our scheme by introducing man-made hybrid attacks. The results demonstrate that our scheme can accurately locate the cropped region. We also show the effectiveness of tamper detection by comparison with some state-of-the-art image forgery detection schemes~\cite{wu2019mantra,kwon2021cat,dong2021mvss}.

The highlights of this paper are three-folded. 
\begin{enumerate}[1)]
\item This paper presents the first high-accuracy robust image crop localization scheme. We innovatively use normalizing flows to build invertible function for image forensic problems. 
\item With the embedded watermark, RWN can also conduct high-accuracy tamper detection, which is comparable with the state-of-the-art works. 
\item We explore a simple yet efficient method (JPEG-Mixup) for improved JPEG simulation. Experiments show that JPEG-Mixup can remarkably improve the generalization of RWN on robustness against JPEG compression. 
\end{enumerate}

\section{Related Works}
\noindent\textbf{Crop Detection and Localization.}
Cropping is a simple yet powerful way to maliciously alter the message of an image. 
The existing crop detection algorithms \cite{fanfani2020vision,yerushalmy2011digital,li2009passive} mainly focus on predicting whether an image is cropped.
For example, Fanfani et al.~\cite{fanfani2020vision} exploits the camera principal point insensitive to image processing operations. 
Yerushalmy et al.~\cite{yerushalmy2011digital} detects whether there are vanishing points and lines on structured image content.
But cropping different areas in a same image will definitely result in different semantic changes. Therefore, we need to predict the position of the crop. 
In \cite{li2009passive}, the block artifact grids (BAGs) are extracted blindly with a new extraction algorithm, and then abnormal BAGs can be detected with a marking procedure as a trail of cropping detection.
For crop localization, Van et al. \cite{van2020dissecting} investigates the impact that cropping has on the image distribution, and predicts the absolute location of image patches. 
However, \cite{van2020dissecting} requires the images to be untouched and uncompressed so that the tiny traces are preserved. 
To address the issue of practical usability, this paper presents RWN for robust image crop localization that does not restrict image format or quality. 

\noindent\textbf{Tamper Localization}
Great efforts have been made on combating daily image forgeries. 
Kown et al.~\cite{kwon2021cat} propose to model quantized DCT coefficient distribution to trace compression artifacts.
Mantra-Net~\cite{wu2019mantra} uses fully convolutional networks for feature extraction and further uses long short-term memory (LSTM) cells for pixel-wise anomaly detection.
In MVSS-Net~\cite{dong2021mvss}, a system with multi-view feature learning and multi-scale supervision is developed to jointly exploit the noise view and the boundary artifact to learn manipulation detection features. In this paper, we study the effectiveness of using watermarking to aid image forensics where we forgo the idea of finding universal traces. 
The shortcoming of these methods is that they only detect limited types of attacks and cannot generalize their performances well on images with a chain of combined attacks.
We robustly hide traces into the original image instead of discovering a universal trace for a much easier forgery localization.

\noindent\textbf{Robust Watermarking. }
The uprising of deep networks have given birth to a series of novel watermarking methods with enhanced robustness~\cite{shin2017jpeg,zhu2018hidden,liu2021jpeg}. 
Shin et al.~\cite{shin2017jpeg} is the first in including a differentiable approximation to JPEG in the watermarking model.
Later, Zhu et al.~\cite{zhu2018hidden} proposes a powerful and comprehensive network that is robust against a variety kinds of attacks.
From then on, subsequent watermarking works~\cite{liu2021jpeg} design more sophisticated and accurate algorithm to simulate image post-processing attacks, especially the JPEG compression.
However, these methods uses fixed quantization table for data compression. In contrast, the quantization table in real-world JPEG images can be customized and flexibly controlled by the quality factor as well as the image content. As a result, the neural networks can over-fit and lack real-world robustness. We explore a simple yet efficient method for improved JPEG simulation.


\section{Method}
\label{section:method}
\subsection{Approach Overview}
\noindent\textbf{Pipeline Design. }
The proposed RWN consists of five stages, namely, watermark embedding, image redistribution, tamper detection, watermark extraction and crop localization. 
We embed a watermark $\mathbf{W}$ into an original image $\mathbf{I}$. The protected image $\mathbf{I}_M$ is generated and we upload it onto the social cloud instead of the unprotected original image. The attacker generates the attacked image $\mathbf{I}_A$ by freely adding three kinds of attacks (benign attacks, cropping, tampering) on $\mathbf{I}_M$. On the recipient's side, the tamper detector $\mathcal{T}$ predicts the tamper mask $\hat{\mathbf{M}}$ on the attacked image to see which parts of the image are tampered, and we also extract the watermark as $\hat{\mathbf{W}}$ from the received image. Afterwards, we rectify the extracted watermark by $\hat{\mathbf{W}}'=\hat{\mathbf{W}}\cdot(1-\hat{\mathbf{M}})$ to discard the tampered contents. Finally, with the original watermark $\mathbf{W}$ as reference, we uses a feature matching algorithm to locate the position of the crop in the original image plane. An image is determined as cropped if there is valid matching result.

\noindent\textbf{Modeling. }
We regard the embedding and extraction as the inverse problem, even if the protected image is cropped and attacked. The formulation of the inverse problem is:
\begin{equation}
  \label{forward}
  (\mathbf{I}_{M}, \mathbf{R}) = \mathcal{P}(\mathbf{I}, \mathbf{W})
\end{equation}
\begin{equation}
  \hat{\mathbf{I}}, \hat{\mathbf{W}} = \mathcal{P}^{-1}(\mathcal{A}(\mathbf{I}_{M}), \hat{\mathbf{R}}),
\end{equation}
subject to $\mathbb{E}(\hat{\mathbf{I}})=\mathbf{I}_{G}, \mathbb{E}(\hat{\mathbf{W}})=\mathbf{W}_{G} \text{ and } \mathbb{E}(\mathbf{G})=\hat{\mathbf{G}}$. 
Here, $\mathbf{R}$ and $\hat{\mathbf{R}}$ are the additional output and input of the network to keep the channel numbers as four. $\mathcal{A}$ denotes the function of image post-processing that attacks the hidden watermark. We let $\hat{\mathbf{R}}=0.5\cdot\mathbbmss{1}$ where $\mathbbmss{1}$ is a matrix full of one. The ground-truth cropped watermark $\mathbf{W}_{G}$ and the ground-truth cropped original image $\mathbf{I}_{G}$ can be generated by sharing the crop mask used by $\mathcal{A}$. 
Besides the invertibility, we regulate that $\mathbb{E}(\mathbf{I}_M)=\mathbf{I}$ for the imperceptibility of the embedding.

\begin{figure}[!t]
	\centering
	\includegraphics[width=0.48\textwidth]{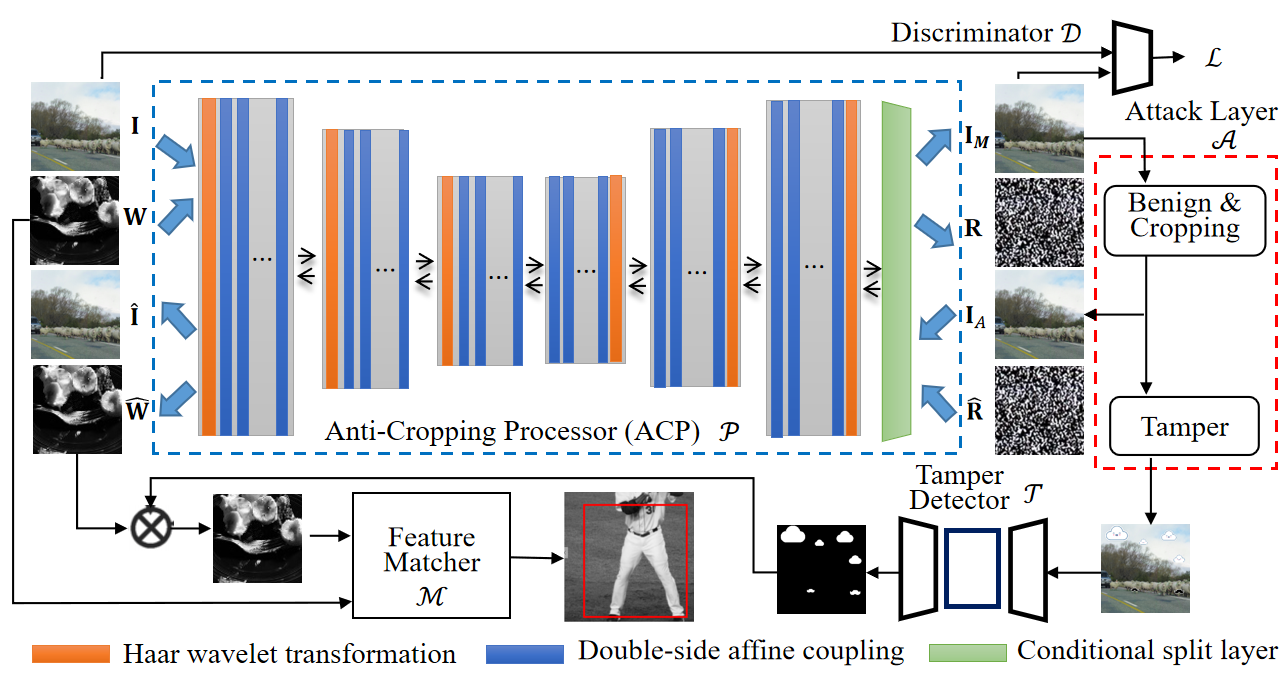}
	\caption{Pipeline design of RWN. The anti-crop processor (ACP) generates the protected image and conversely in the back-propagation recovers the hidden watermark. The attack layer simulates several kinds of digital attacks. The tamper detector predicts the tamper mask and the feature detector locates the cropped region of the attacked image.}
	\label{fig_framework}
\end{figure}

Fig. \ref{fig_framework} shows the pipeline of our scheme. We employ an Anti-Cropping Processor (ACP) $\mathcal{P}$ that jointly learns the paired function of both watermark embedding and extraction. We implement $\mathcal{A}$ using a differentiable attack layer that simulates the attacker's behavior. The rest of the components include a tamper detector $\mathcal{T}$, a discriminator $\mathcal{D}$ and a feature matcher $\mathcal{M}$, which is the SURF algorithm.


\begin{figure}[!t]
	\centering
	\includegraphics[width=0.5\textwidth]{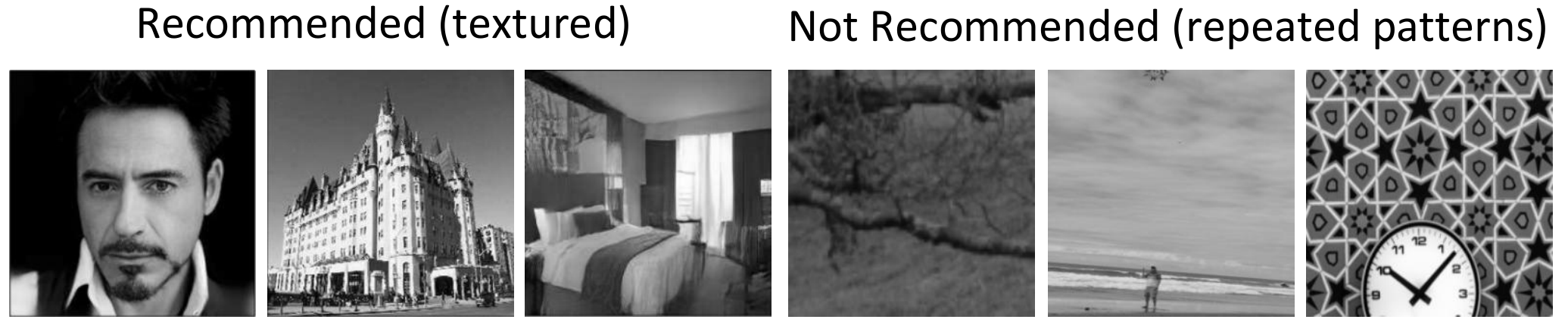}
	\caption{Watermark selection. We choose textured watermarks (left) instead of ones with repeated patterns (right).}
	\label{fig_repeat}
\end{figure}
\begin{figure*}[!t]
	\centering
	\includegraphics[width=1.0\textwidth]{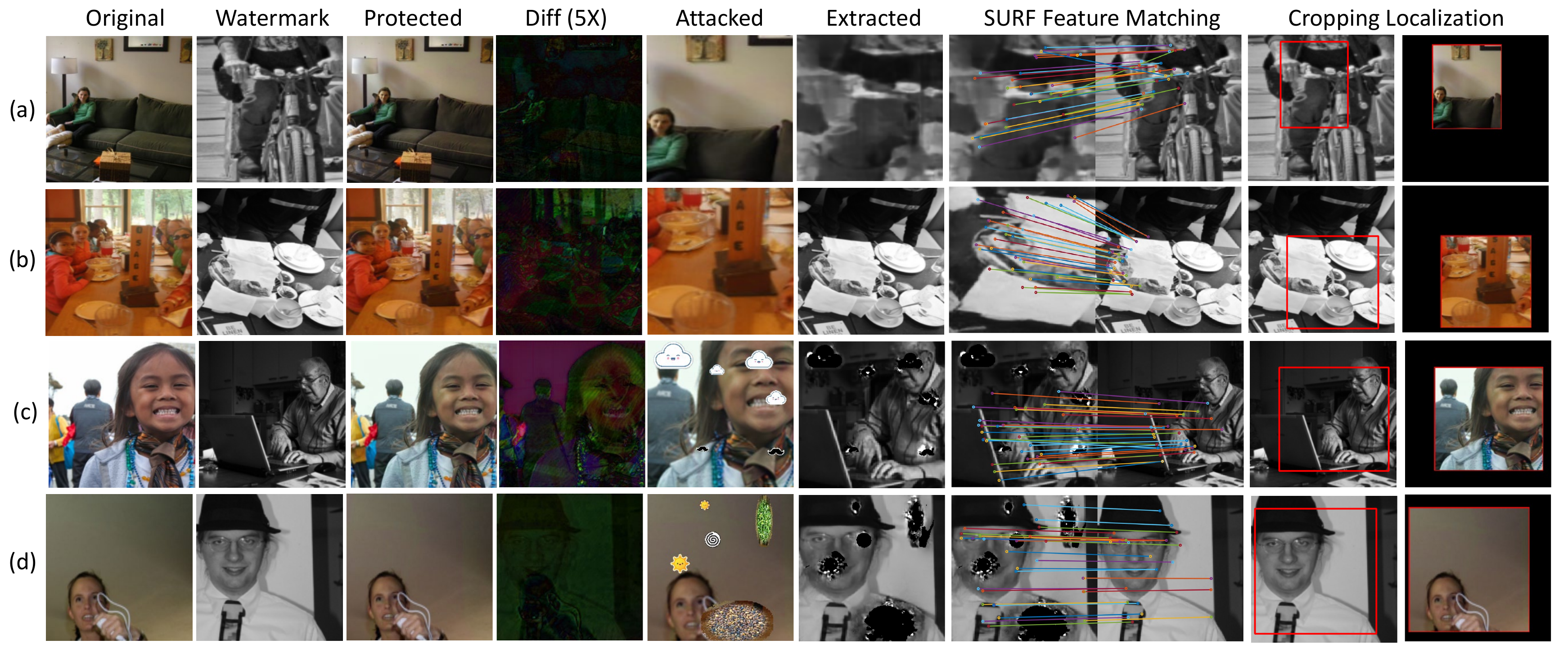}
	\caption{Results of watermark embedding, extraction and crop localization. The attacks involved in each group are different, namely, (a) crop alone, (b) crop \& scaling, (c) crop \& JPEG \& tamper and (d) crop \& Median blurring \& tamper.}
	\label{img_marked_extract}
\end{figure*}
\subsection{Network Implementation}
Considering the efficiency of the invertible U-Net proposed in~\cite{xiao2020invertible}, we build our ACP on top of this architecture. The network consists of six invertible blocks each of which contains a Haar wavelet transformation and a double-side affine coupling. The network ends with a conditional split layer. The number of the input and output channels are four. The sizes of $\mathbf{R}$ and $\hat{\mathbf{R}}$ are of the same as $\mathbf{W}$. $\mathbf{R}$ and $\hat{\mathbf{R}}$ are not required to be the same.

In the attacking layer $\mathcal{A}$, we first convert the marked images into the corresponding 8-bit RGB-formatted image. We adopt the Straight-Through Estimator~\cite{bengio2013estimating} on calculating the gradients. 
Then, we build differentiable methods to simulate the hybrid cropping attacks, which include benign attacks, cropping and tampering attack. The benign attacks are represented by JPEG compression, scaling, etc., which do not alter the semantic meaning of the image. We take the implementation from~\cite{zhu2018hidden} except that we build our own JPEG simulator. For cropping, we randomly crop a portion of the protected image $\textbf{I}_M$. For tampering attack, we first randomly select random areas using a binary matrix $\mathbf{M}$ inside $\mathbf{I}_M$, and generate the tampered image by 
\begin{equation}
\mathbf{I}_{A}=\mathbf{I}_{\emph{irr}} \cdot \mathbf{M}+\mathbf{I}_{M} \cdot (1-\mathbf{M}_{R}),
\end{equation}
where $\mathbf{I}_{\emph{irr}}$ refers to the source of the tamper.

Although there are already many scheme which include a carefully-designed JPEG simulator, e.g., JPEG-SS~\cite{liu2021jpeg}, JPEG-Mask~\cite{zhu2018hidden}, MBRS~\cite{jia2021mbrs}, the real-world JPEG robustness of these schemes is still limited. We believe it mainly attribute to that the networks are over-fitted to a fixed compression mode.
In this paper, we propose to apply the Mix-Up strategy~\cite{zhang2017mixup} over these implementations so that the generated results are more flexible. 
We randomly sample a quality factor $QF_{1}$ that follows $QF_{1}\in[10,100]$, together with a previous differentiable implementation of JPEG simulator $\mathcal{J}_{1}$, where $\mathcal{J}_{1}\in $\{JPEG-SS, JPEG-Mask, MBRS\}. We correspondingly generate a pseudo JPEG image $\mathbf{I}_{\emph{jpg}}^{1}$ by $\mathbf{I}_{\emph{jpg}}^{1}=\mathcal{J}_{1}(\mathbf{I}_{M},QF_{1})$.
In the same fashion, we generate $k-1$ more pseudo-JPEG images, denoted as $\mathbf{I}_{\emph{jpg}}^{2}, ..., \mathbf{I}_{\emph{jpg}}^{k}$.
Finally, the output of JPEG-Mixup is produced by mixing the pseudo-JPEG images with arbitrary contributing rates $\epsilon_{l}$.
\begin{equation}
\mathbf{I}_{jpg}=\sum_{{l}\in [1,k]}\epsilon_{l}\cdot\mathcal{J}_{l}(\mathbf{I}_{M}, QF_{l}).
\label{eqn_JPEG}
\end{equation}
where $\sum\epsilon_{l}=1$.
The proposed mix-up based JPEG simulation (JPEG-Mixup) is fairly simple yet effective. Compared to previous schemes, the interpolation technique can generates more diversed pseudo-JPEG images for RWN.
In the experiments, we show the performance gain in JPEG robustness using JPEG-Mixup in comparison with the other simulators.

Finally, we build the tamper detector $\mathcal{T}$ upon U-Net~\cite{ronneberger2015u}, a traditional image segmentation network, and implement the discriminator $\mathcal{D}$ using Patch-GAN~\cite{isola2017image}. 

\noindent\textbf{Watermark Selection.} We choose gray-scaled natural images with rich texture as valid watermark. Images with too much repetitive patterns will lead to multiple matching solutions for the SURF detector. Also, considering that the recipient can only blindly extract a partial watermark, we use those common images available on the Internet that can be easily downloaded by the recipient using image retrieval systems by referring to the extracted watermark. Fig.~\ref{fig_repeat} shows some examples of recommended/not recommended watermarks.

\subsection{Objective Loss Function and Training Details}
\label{subsection:lossfunc}
For ACP, the first part of the loss is reconstruction loss $\mathcal{L}_{\emph{rec}}$ for $\mathbb{E}(\hat{\mathbf{W}})=\mathbf{W}_G$, $\mathbb{E}(\hat{\mathbf{I}})=\mathbf{I}_G$ and $\mathbb{E}(\mathbf{I}_M)=\mathbf{I}$.
$\mathbb{E}(\cdot)$ is the expectation operator.
$\mathcal{L}_{\emph{rec}}=\mathcal{F}(\mathbf{I}, \mathbf{I}_M)+\mathcal{F}(\mathbf{W}_{G}, \hat{\mathbf{W}})+\mathcal{F}(\mathbf{I}_{G}, \hat{\mathbf{I}})$, where $\mathcal{F}$ is the $\ell_1$ distance function. The second part is the nullification loss for $\mathbb{E}(\mathbf{R})=\hat{\mathbf{R}}$. $\mathcal{L}_{\emph{ran}}= \mathcal{F}(\mathbf{R}, \hat{\mathbf{R}})$. The third part is the adversarial loss, which further control the introduced distortion by fooling the discriminator. We accept the least squared adversarial loss (LS-GAN)~\cite{mao2017least}. The total loss for ACP is $
\mathcal{L}_{\mathcal{G}}=\mathcal{L}_{rec}+\alpha\cdot\mathcal{L}_{dis}+\beta\cdot\mathcal{L}_{ran}$, where $\alpha$ and $\beta$ are hyper-parameters. For the tamper detector, we minimize the binary cross entropy (BCE) loss between the predicted tamper mask $\hat{\mathbf{M}}$ and the ground-truth mask $\mathbf{M}$. $
\mathcal{L}_{\mathcal{T}}=BCE(\mathbf{M},\hat{\mathbf{M}}).$

We resize the images to the size of $256\times~256$. The hyper-parameters are set as $\alpha=1, \beta=8$. We set $k$ as 3, which is the generated pseudo-JPEG images used in JPEG-Mixup. The batch size is set as 16, and we empirically find that using Batch Normalization (BN) in our scheme outperforms using Instance Normalization (IN).
We use Adam optimizer~\cite{kingma2014adam} with the default parameters. The learning rate is $1\times10^{-4}$.




\section{Experimental Results}
\label{section:experiment}
\subsection{Experiment Setup}
We train the scheme on the COCO training/test set~\cite{lin2014microsoft} with automatically generated attacks. The dataset contains 117266 training samples and 40670 testing samples. The scheme is tested with human-participated attacks, where we have invited some volunteers to crop the provided marked images and perform further image post-processing attack on them. The crop rate is roughly $\delta\in[0.5,1)$. During testing, we use real JPEG coder to compress the protected images.

\noindent\textbf{Evaluation Metric. }
We employ peak signal-to-noise ratio (PSNR), structural similarity (SSIM) \cite{wang2004image} to evaluate the image quality, and F1 score \cite{ren2015faster} to measure the accuracy of crop/tamper localization. 

\noindent\textbf{Benchmark.}
We compare RWN with Van et al.~\cite{van2020dissecting} for crop localization. 
In addition, we compare RWN with Mantra-Net~\cite{wu2019mantra}, MVSS-Net~\cite{dong2021mvss} and CAT-Net~\cite{kwon2021cat} for tamper localization. 

\begin{figure*}[!t]
	\centering
	\includegraphics[width=1.0\textwidth]{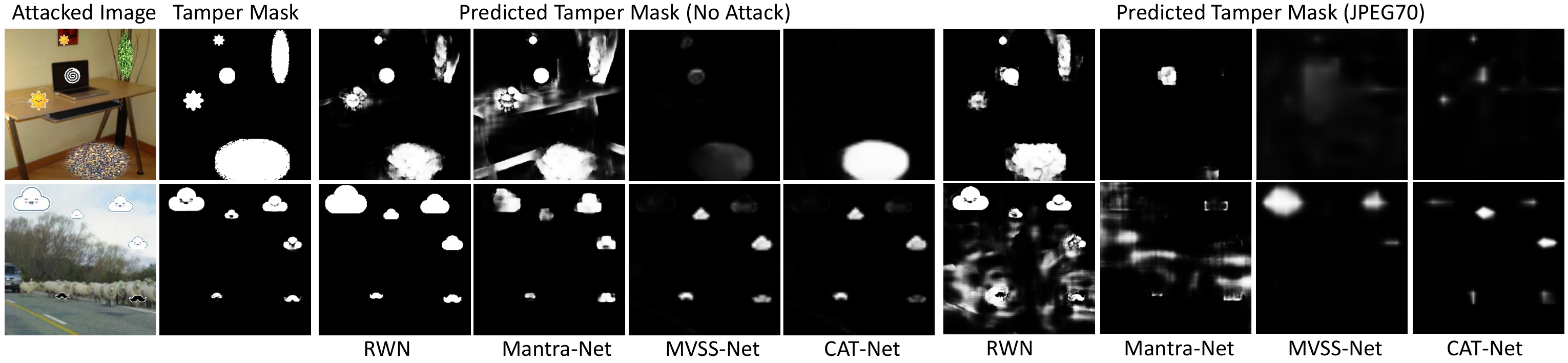}
	\caption{We compare the result of tamper localization on two test images with several state-of-the-art image forgery detection schemes.}
	\label{img_locate}
\end{figure*}
\subsection{Real-World Performance of Crop Localization}
\noindent\textbf{Quality of the Protected Images.} In Fig.~\ref{img_marked_extract}, we randomly sample different pairs of images as the original and watermark. From the figures, generally, little detail of the watermark can be found. Though the magnified difference is visible, it does not matter if the difference is visible, only that the marked image is perceptually close to the original image.
We have conducted more embedding experiments over 1000 images from the test set, and the average PSNR between the protected images and the original images is 36.23dB, and the average SSIM~\cite{wang2004image} is 0.983. 

\noindent\textbf{Accuracy of the Crop Localization.} Fig.~\ref{img_marked_extract} further shows the results of watermark extraction, feature matching and crop localization. In row (a), we only crop the protected image. We see that the crop mask is accurately predicted. As a result, even without the prior knowledge of the original image, we know the relative position of the attacked image in the original image plane. We have conducted more experiments over 1000 images under different crop rate and different kinds of benign attacks. Here for space limit, we take $QF$=70 as an example. The average performances are reported in Table~\ref{table_crop}. Higher IoU (Intersection over Union) and SSIM~\cite{wang2004image} indicates more accurate localization result.  The average IoUs are above 0.8 where the protected images do not undergo further attacks except cropping. The scheme is proven to be agnostic to the crop size in that the performance does not degrade significantly with larger crop rate.

\begin{table}[!t]
	\caption{Average performance of cropping localization under different cropping rate and attack.}
	\small
    \setlength{\tabcolsep}{3pt}
	\label{table_crop}
	\begin{center}
	\begin{tabular}{c|c|c|c|c|c|c}
	\hline
		Rate & Index & NoAttack & JPEG & Scaling & MedianBlur & AWGN \\
		\hline
		\multirow{2}{*}{90\%} & IoU & 0.919 & 0.895  & 0.803 & 0.838 & 0.784\\
		& SSIM & 0.949  & 0.896  & 0.957 & 0.953 & 0.832\\
		\hline
		\multirow{2}{*}{70\%} & IoU & 0.858 & 0.813  & 0.878 & 0.915 & 0.763\\
		& SSIM & 0.942 & 0.878 & 0.940 & 0.940 & 0.804 \\
		\hline
		\multirow{2}{*}{50\%} & IoU & 0.821 & 0.706  & 0.603 & 0.540 & 0.513\\
		 & SSIM & 0.914 & 0.7223 & 0.942 & 0.840 & 0.788\\
		\hline
	\end{tabular}
	\end{center}
\end{table}
\begin{table}[!t]
	\caption{F1 score comparison for tamper detection among our scheme and the state-of-the-art methods.}
	\label{table_localization}
	\small
    \setlength{\tabcolsep}{3pt}
    \begin{center}
    \begin{tabular}{c|c|c|c|c|c}
    \hline
    		Method & NoAttack & JPEG & Blurring & Scaling & AWGN\\
    		\hline
    		Proposed & 0.773 & 0.736  & 0.695 & 0.745 & 0.573\\
    		\hline
    		Mantra-Net~\cite{wu2019mantra}& 0.566 & 0.480 & 0.557 & 0.540 & 0.347\\
    		\hline
			MVSS-Net~\cite{dong2021mvss}& 0.545 & 0.364 & 0.399 & 0.485 & 0.323\\
    		\hline
    		CAT-Net~\cite{kwon2021cat}& 0.467 & 0.433  & 0.419 & 0.428 & 0.365\\
    		\hline
    	\end{tabular}
    	\end{center}
\end{table}
\noindent\textbf{Robustness.}
We test the robustness of our scheme by conducting image post-processing attack on the marked image. In the last two groups, tampering attack is introduced. In Table~\ref{table_crop}, we can observe that our scheme provides high-accuracy crop localization despite the presence of the attacks along with cropping. The results promote the real-world application of RWN. Thanks to the robustness of feature matching of SURF, the experiments show that in most cases our scheme do not require a precise watermark extraction.

\noindent\textbf{Comparison. } 
Van et al.~\cite{van2020dissecting} is fragile where the targeted image cannot be compressed or low-resolutioned. If the original images are in JPEG formats, the required traces are much likely destroyed already.
The F1 score of \cite{van2020dissecting} are 0.575 on natural images where the resolution and image quality are randomized. In contrast, RWN does not have any restriction on the original image. After watermarking, we can robustly locate the crop with a much higher F1 score 0.867.

\subsection{Accuracy of Tamper Detection.}
Fig.~\ref{img_locate} shows the tamper detection results on two test images. Note that for fair comparison, we add the same attacks on the original images for \cite{wu2019mantra,kwon2021cat,dong2021mvss}. We observe that although the images are tampered by a variety of random attacks, we succeed in localizing the tampered areas. In Table~\ref{table_localization}, we provide the average results over 1000 images. The F1 score of RWN is 0.773 on uncompressed images and 0.736 on JPEG attacked images ($QF=80$). The performance of Mantra-Net on JPEG images is much worse than that on plain-text images. We believe the reason is that less statistical trace is preserved in the compressed version. In contrast, the embedded watermark signal serves as the alternative trace for tamper detection, which is designed to resist benign attacks. The performance on JPEG images does not drop too much. 
Therefore, robust watermarking can successfully aid tamper detection by hiding crafted traces similar to~\cite{yin2018deep,khachaturov2021markpainting}.

\begin{figure}[!t]
	\centering
	\includegraphics[width=0.49\textwidth]{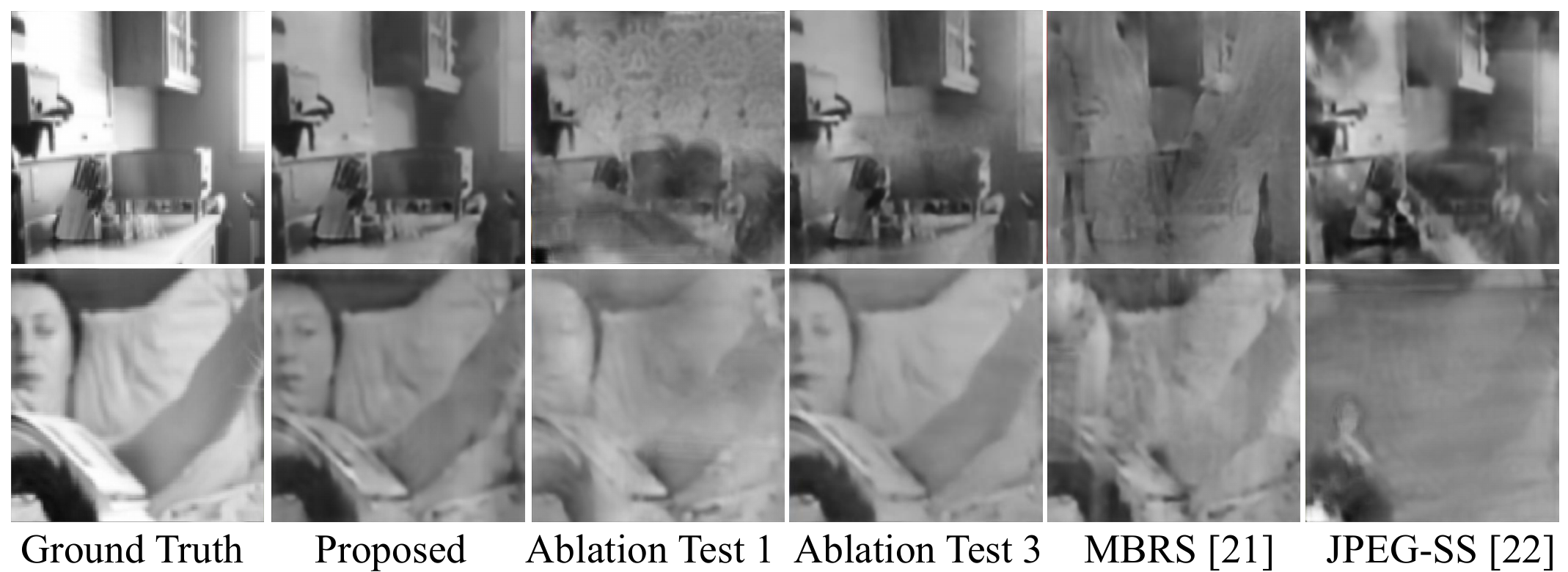}
	\caption{Results of the ablation studies where we validate the network design and the Mix-Up~\cite{zhang2017mixup} strategy in JPEG simulation.}
	\label{fig_ablation}
\end{figure}
\subsection{Ablation Study}
\label{section_ablation}
We discuss the influences of the network design and the training strategies in RWN. In each ablation test, we fine-tune the network till the accuracy of cropping localization is close to the baseline. We also use fixed attacks during testing. The experiments are conducted as follows.

\noindent\textbf{Influence of INN architecture.} In Test 1, we train two individual fully convolutional networks to implement the hiding network and revealing network. The normalizing-flow-based ACP is replaced in order to measure the effectiveness brought by invertible problem formulation. 

\noindent\textbf{Influence of JPEG Simulator.} In Test 2, we implement the JPEG attack with that proposed in~\cite{jia2021mbrs,zhu2018hidden,liu2021jpeg}. The JPEG QF and the crop ratio are kept the same for fair comparison. We train the implementations together with the baseline under the same losses and batch size. 

\noindent\textbf{Influence of Discriminator.} In Test 3, we do not use the discriminator $\mathcal{D}$ and train the pipeline with only the reconstruction and the BCE loss.

Fig.~\ref{fig_ablation} shows the detailed comparison results. First, the baseline results outperform the encoder-decoder network design. Second, while MBRS~\cite{jia2021mbrs} can provide decent robustness, the extraction performance is even better using our JPEG simulator. Specifically, the average SSIM between $\hat{\mathbf{W}}$ and $\mathbf{W}_{G}$ using MBRS~\cite{jia2021mbrs} is 0.797 compared to 0.878 reported in Table.\ref{table_crop}. The Mix-Up strategy prevents the networks from being over-fitted to any single JPEG simulator, which helps the scheme significantly improve its real-world robustness. Third, without the discriminator, the extraction result is also worse than the baseline.


\section{Conclusion}
This paper presents a novel robust watermarking network for image crop localization. 
We use an invertible pipeline for watermark embedding and extraction, where the watermark is shared with the recipient. ACP produces a visually indistinguishable protected image, which is then posted on the social network instead of the unprotected version. On receiving the attacked version of the protected image, we use ACP again to extract the watermark, and conduct SURF feature matching algorithms to 
determine where the crop was positioned in the original image plane. 
We also extend our scheme to detect tampering attack on the attacked image.
Experiments verify that RWN is effective in real-world application.
Finally, we prove the necessity of the progressive recovery as well as the network design.
\bibliographystyle{IEEEbib}
\bibliography{reference}

\begin{thebibliography}{10}

\bibitem{wu2019mantra}
Yue Wu, Wael AbdAlmageed, and Premkumar Natarajan,
\newblock ``Mantra-net: Manipulation tracing network for detection and
  localization of image forgeries with anomalous features,''
\newblock in {\em Proceedings of the IEEE/CVF Conference on Computer Vision and
  Pattern Recognition}, 2019, pp. 9543--9552.

\bibitem{kwon2021cat}
Myung-Joon Kwon, In-Jae Yu, Seung-Hun Nam, and Heung-Kyu Lee,
\newblock ``Cat-net: Compression artifact tracing network for detection and
  localization of image splicing,''
\newblock in {\em Proceedings of the IEEE/CVF Winter Conference on Applications
  of Computer Vision}, 2021, pp. 375--384.

\bibitem{dong2021mvss}
Xinru Chen, Chengbo Dong, Jiaqi Ji, Juan Cao, and Xirong Li,
\newblock ``Image manipulation detection by multi-view multi-scale
  supervision,''
\newblock in {\em Proceedings of the IEEE/CVF International Conference on
  Computer Vision}, 2021, pp. 14185--14193.

\bibitem{li2020face}
Lingzhi Li, Jianmin Bao, Ting Zhang, Hao Yang, Dong Chen, Fang Wen, and Baining
  Guo,
\newblock ``Face x-ray for more general face forgery detection,''
\newblock in {\em Proceedings of the IEEE/CVF Conference on Computer Vision and
  Pattern Recognition}, 2020, pp. 5001--5010.

\bibitem{fanfani2020vision}
Marco Fanfani, Massimo Iuliani, Fabio Bellavia, Carlo Colombo, and Alessandro
  Piva,
\newblock ``A vision-based fully automated approach to robust image cropping
  detection,''
\newblock {\em Signal Processing: Image Communication}, vol. 80, pp. 115629,
  2020.

\bibitem{yerushalmy2011digital}
Ido Yerushalmy and Hagit Hel-Or,
\newblock ``Digital image forgery detection based on lens and sensor
  aberration,''
\newblock {\em International journal of computer vision}, vol. 92, no. 1, pp.
  71--91, 2011.

\bibitem{li2009passive}
Weihai Li, Yuan Yuan, and Nenghai Yu,
\newblock ``Passive detection of doctored jpeg image via block artifact grid
  extraction,''
\newblock {\em Signal Processing}, vol. 89, no. 9, pp. 1821--1829, 2009.

\bibitem{van2020dissecting}
Basile Van~Hoorick and Carl Vondrick,
\newblock ``Dissecting image crops,''
\newblock in {\em Proceedings of the IEEE/CVF International Conference on
  Computer Vision}, 2021, pp. 9741--9750.

\bibitem{ying2019robust}
Qichao Ying, Jingzhi Lin, Zhenxing Qian, Haisheng Xu, and Xinpeng Zhang,
\newblock ``Robust digital watermarking for color images in combined dft and
  dt-cwt domains,''
\newblock {\em Mathematical Biosciences and Engineering}, vol. 16, no. 5, pp.
  4788--4801, 2019.

\bibitem{zhu2018hidden}
Jiren Zhu, Russell Kaplan, Justin Johnson, and Li~Fei-Fei,
\newblock ``Hidden: Hiding data with deep networks,''
\newblock in {\em Proceedings of the European conference on computer vision
  (ECCV)}, 2018, pp. 657--672.

\bibitem{mun2019finding}
Seung-Min Mun, Seung-Hun Nam, Haneol Jang, Dongkyu Kim, and Heung-Kyu Lee,
\newblock ``Finding robust domain from attacks: A learning framework for blind
  watermarking,''
\newblock {\em Neurocomputing}, vol. 337, pp. 191--202, 2019.

\bibitem{khachaturov2021markpainting}
David Khachaturov, Ilia Shumailov, Yiren Zhao, Nicolas Papernot, and Ross
  Anderson,
\newblock ``Markpainting: Adversarial machine learning meets inpainting,''
\newblock in {\em International Conference on Machine Learning}. PMLR, 2021,
  pp. 5409--5419.

\bibitem{yin2018deep}
Minghao Yin, Yongbing Zhang, Xiu Li, and Shiqi Wang,
\newblock ``When deep fool meets deep prior: Adversarial attack on
  super-resolution network,''
\newblock in {\em Proceedings of the 26th ACM international conference on
  Multimedia}, 2018, pp. 1930--1938.

\bibitem{kingma2018glow}
Durk~P Kingma and Prafulla Dhariwal,
\newblock ``Glow: Generative flow with invertible 1x1 convolutions,''
\newblock {\em Advances in neural information processing systems}, vol. 31,
  2018.

\bibitem{bay2006surf}
Herbert Bay, Tinne Tuytelaars, and Luc Van~Gool,
\newblock ``Surf: Speeded up robust features,''
\newblock in {\em European conference on computer vision}. Springer, 2006, pp.
  404--417.

\bibitem{shin2017jpeg}
Richard Shin and Dawn Song,
\newblock ``Jpeg-resistant adversarial images,''
\newblock in {\em NIPS 2017 Workshop on Machine Learning and Computer
  Security}, 2017, vol.~1.

\bibitem{liu2021jpeg}
Kunlin Liu, Dongdong Chen, Jing Liao, Weiming Zhang, Hang Zhou, Jie Zhang,
  Wenbo Zhou, and Nenghai Yu,
\newblock ``Jpeg robust invertible grayscale,''
\newblock {\em IEEE Transactions on Visualization and Computer Graphics}, 2021.

\bibitem{xiao2020invertible}
Mingqing Xiao, Shuxin Zheng, Chang Liu, Yaolong Wang, Di~He, Guolin Ke, Jiang
  Bian, Zhouchen Lin, and Tie-Yan Liu,
\newblock ``Invertible image rescaling,''
\newblock in {\em European Conference on Computer Vision}. Springer, 2020, pp.
  126--144.

\bibitem{bengio2013estimating}
Yoshua Bengio, Nicholas L{\'e}onard, and Aaron Courville,
\newblock ``Estimating or propagating gradients through stochastic neurons for
  conditional computation,''
\newblock {\em arXiv preprint arXiv:1308.3432}, 2013.

\bibitem{jia2021mbrs}
Zhaoyang Jia, Han Fang, and Weiming Zhang,
\newblock ``Mbrs: Enhancing robustness of dnn-based watermarking by mini-batch
  of real and simulated jpeg compression,''
\newblock in {\em Proceedings of the 29th ACM International Conference on
  Multimedia}, 2021, pp. 41--49.

\bibitem{zhang2017mixup}
Hongyi Zhang, Moustapha Cisse, Yann~N Dauphin, and David Lopez-Paz,
\newblock ``mixup: Beyond empirical risk minimization,''
\newblock {\em arXiv preprint arXiv:1710.09412}, 2017.

\bibitem{ronneberger2015u}
Olaf Ronneberger, Philipp Fischer, and Thomas Brox,
\newblock ``U-net: Convolutional networks for biomedical image segmentation,''
\newblock in {\em International Conference on Medical image computing and
  computer-assisted intervention}. Springer, 2015, pp. 234--241.

\bibitem{isola2017image}
Phillip Isola, Jun-Yan Zhu, Tinghui Zhou, and Alexei~A Efros,
\newblock ``Image-to-image translation with conditional adversarial networks,''
\newblock in {\em Proceedings of the IEEE conference on computer vision and
  pattern recognition}, 2017, pp. 1125--1134.

\bibitem{mao2017least}
Xudong Mao, Qing Li, Haoran Xie, Raymond~YK Lau, Zhen Wang, and Stephen
  Paul~Smolley,
\newblock ``Least squares generative adversarial networks,''
\newblock in {\em Proceedings of the IEEE international conference on computer
  vision}, 2017, pp. 2794--2802.

\bibitem{kingma2014adam}
Diederik~P Kingma and Jimmy Ba,
\newblock ``Adam: A method for stochastic optimization,''
\newblock {\em arXiv preprint arXiv:1412.6980}, 2014.

\bibitem{lin2014microsoft}
Tsung-Yi Lin, Michael Maire, Serge Belongie, James Hays, Pietro Perona, Deva
  Ramanan, Piotr Doll{\'a}r, and C~Lawrence Zitnick,
\newblock ``Microsoft coco: Common objects in context,''
\newblock in {\em European conference on computer vision}. Springer, 2014, pp.
  740--755.

\bibitem{wang2004image}
Zhou Wang, Alan~C Bovik, Hamid~R Sheikh, and Eero~P Simoncelli,
\newblock ``Image quality assessment: from error visibility to structural
  similarity,''
\newblock {\em IEEE transactions on image processing}, vol. 13, no. 4, pp.
  600--612, 2004.

\bibitem{ren2015faster}
Shaoqing Ren, Kaiming He, Ross Girshick, and Jian Sun,
\newblock ``Faster r-cnn: Towards real-time object detection with region
  proposal networks,''
\newblock {\em Advances in neural information processing systems}, vol. 28, pp.
  91--99, 2015.

\end{thebibliography}

\end{document}